\documentclass[journal,twoside,web]{ieeecolor}
\usepackage{generic}
\usepackage{cite}
\usepackage{amsmath,amssymb,amsfonts}
\usepackage{algorithmic}
\usepackage{booktabs}
\usepackage{fixmath}
\usepackage{graphicx}
\usepackage{textcomp}
\usepackage{mathtools}
\usepackage[export]{adjustbox}
\usepackage{float}
\usepackage{multirow}
\usepackage[hidelinks]{hyperref}
\usepackage[flushleft]{threeparttable}
\usepackage[ruled,vlined]{algorithm2e}
\usepackage{wrapfig}
\usepackage{pifont}

\usepackage{bm}
\usepackage{subcaption}
\usepackage[flushleft]{threeparttable} %
\usepackage{booktabs}
\usepackage[dvipsnames]{xcolor}

\definecolor{darkgreen}{RGB}{0,120,0}

\def\BibTeX{{\rm B\kern-.05em{\sc i\kern-.025em b}\kern-.08em
		T\kern-.1667em\lower.7ex\hbox{E}\kern-.125emX}}
\begin{document}
	\title{
        Temporal Cardiovascular Dynamics for Improved PPG-Based Heart Rate Estimation
	}
	\author{Berken Utku Demirel and Christian Holz
		\thanks{Received 7 February 2025; revised 21 July 2025; accepted 25
September 2025.
        The authors are with the Department of Computer Science, ETH Zürich, Stampfenbachstrasse 48, 8092 Zurich, Switzerland (e-mail: berken.demirel@inf.ethz.ch, christian.holz@inf.ethz.ch).}}
	
	\maketitle
	
\begin{abstract}
The oscillations of the human heart rate are inherently complex and non-linear---they are best described by \textit{mathematical chaos}, and they present a challenge when applied to the practical domain of cardiovascular health monitoring in everyday life.
In this work, we study the non-linear chaotic behavior of heart rate through mutual information and introduce a novel approach for enhancing heart rate estimation in real-life conditions.
Our proposed approach not only explains and handles the non-linear temporal complexity from a mathematical perspective but also improves the deep learning solutions when combined with them.
We validate our proposed method on four established datasets from real-life scenarios and compare its performance with existing algorithms thoroughly with extensive ablation experiments. 
Our results demonstrate a substantial improvement, up to 40\%, of the proposed approach in estimating heart rate compared to traditional methods and existing machine-learning techniques while reducing the reliance on multiple sensing modalities and eliminating the need for post-processing steps.

\end{abstract}
	

\section{Introduction}
\label{sec:introduction}


Healthy biological systems exhibit complex patterns of variability that can be described by \textit{mathematical chaos}~\cite{chaos_is_a_ladder,complex_hr}.
A healthy heart is not a metronome; instead, its complex and constantly changing oscillations enable the cardiovascular system to rapidly adjust to sudden physical and psychological challenges to homeostasis~\cite{complex_hr}.
Therefore, measuring heart rate (HR) during daily life has significant importance in monitoring individuals' health.
For example, increased oscillations of HR measurements are related to cardiac conduction abnormalities, which are strongly linked to an increased risk of mortality (particularly among the elderly)~\cite{HRV}.

Moreover, the changes in heart rate to activities, such as HR variations during running, have been shown to be predictive of cardiovascular disease (CVD) risk and are inversely associated with many other health outcomes~\cite{JAMA_hrv}.
Therefore, monitoring heart rate variations in daily life, espeically during activities, not only provides valuable insights into cardiovascular health but also serves as a predictive tool for assessing the risk of cardiovascular diseases and cardiovascular mortality~\cite{sandvik_heart_1995}.

The electrocardiogram (ECG) is the primary tool for monitoring cardiac activity.
However, it requires electrodes to be connected to various parts of the body and, therefore, can only be used for a limited amount of time within medical facilities~\cite{limited_ecg}.
Recently, wearable devices have emerged as powerful platforms for monitoring the cardiac activity of users during their daily life through blood flow measurement sensors, i.e.,  photoplethysmogram (PPG). 
Commercial products, such as the Apple Watch~\cite{Apple_watch}, Google's Fitbit~\cite{Fitbit}, or the Samsung Watch~\cite{HRV_UCI}, now include these sensors to track the wearer's heart rate, for health monitoring and fitness purposes.

However, measurements on wearable devices are prone to motion artifacts (MAs), which cause variability of sensor pressure on the skin while scattering or leaking ambient light into the gap between the photodiode and the skin, resulting in a significant decrease in heart rate estimation performance~\cite{HRV_UCI}. 

For instance, when users engage in physical activities, such as walking, running, or working, the HR estimation performance in wearable devices decreases heavily. 
The studies that are concerned with commercial wearables state that the performance of devices extremely decreases during high-intensity activities.
For example, reviewing studies on Fitbit reported severe underestimations of the HR\textcolor{black}{\cite{poor_fitbit,poor_fitbid_2, Shannon_PPG}}.
Similarly, previous works showed that the performance of the Apple Watch for HR estimation during exercise decreases with increasing intensity and the proportion of HR values recorded by the watch decreases~\cite{poor_apple_watch}.
In other words, previous works reported a trivial underestimation of the heart rate from the Apple Watch compared to the ground truth~\cite{poor_apple_2}.

\begin{figure}[t]
    \centering
    \includegraphics[width=0.9\columnwidth]{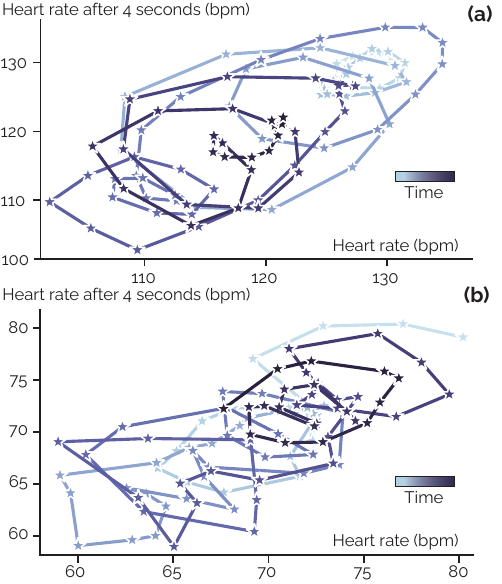}
    \caption{Heart rate dynamics during walking \textbf{(a)} and reading \textbf{(b)}. 
    Both figures show the phase space plot of sinus rhythm for two healthy subjects over three minutes.
    Each point represents the current HR (x-axis) and the HR after 4 seconds (y-axis). 
    The gradual darkening of the line color represents the time, which illustrates the chaotic fluctuations of heart rate.}
    \label{fig:enter-label}
\end{figure}

Moreover, a recent study shows that in a commercial device, HR estimation errors reach 80--100 beats per minute (bpm) during wake periods where data recorded from sensors are contaminated by different levels of noise~\cite{HRV_UCI}.
While the Association for the Advancement of Medical Instrumentation (AAMI) standard indicates that for cardiac monitors the allowable error rate should be less than $\pm 5$\,bpm or $\pm 10\%$, whichever is greater~\cite{ansi_aami}. 
Although this error is defined for a single estimation instead of the mean, current state-of-the-art methods, especially those using deep learning, fall short of this performance.
Moreover, estimating the HR during intensive activities is crucially important for monitoring individuals' cardiovascular health while the current solutions fail.
Current solutions for estimating HR from blood volume pulse signals remain straightforward while applying machine learning as they have not dealt with the complex and non-linear dynamics of the cardiovascular system, which limits performance and overshadows the learning capability of algorithms.

In this paper, we first study the chaotic behavior of the heart rate in real-life conditions through mutual information and introduce a novel approach for enhancing heart rate estimation with deep learning solutions.
Our proposed method overcomes previous limitations of motion artifacts by learning the chaotic behavior of cardiovascular dynamics together with the sensor data and utilizes this for predictions during inference without requiring any additional post-processing or sensor modalities. 
The contributions can be summarized as follows:

\begin{itemize}
    \item We present an information-theoretical approach for understanding the non-linear chaotic behavior of cardiovascular dynamics.
    To the best of our knowledge, we are the first to investigate the non-linear temporal relations in the heart rate with a mathematical approach to identify underlying patterns within this complex behavior, shedding light on previously unrecognized order.

    \item Building on this, we integrate our findings into deep learning solutions to improve the heart rate estimations, different from intuition-based traditional methods and trial-error-based learning solutions.

    \item We extensively evaluated our method on established HR estimation datasets that span a wide range of physical activities, performed under close to real-life conditions.
    Our method outperforms the prior works in HR estimation.
    Unlike previous methods, our method produces estimations with high robustness over time while decreasing the number of required sensing modalities and discarding the need for post-processing.
\end{itemize}



\section{Related Works}
\label{sec:Related_works}
A considerable amount of literature has been published on estimating heart rate from wearable devices based on PPG signals.
Most of these studies utilized power spectral density (PSD) estimation, a quantitative method for describing a signal in terms of its underlying oscillations, enabling one to observe how the frequency changes over time.
Assuming the heart rate will not change abruptly and be stationary in a short window, the PSD reveals information about heartbeat frequency.
While this approach works for less contaminated PPG, motion artifacts hinder the measurement of heart rate in spectral density. 
As a result, different approaches have been proposed to obtain the PSD of PPG signals with accelerometers to differentiate the frequency of heartbeats from motion artifacts~\cite{DeepPPG,sch_original,spa_original, phase_voco_ppg, JBHI_FSM2, JBHI_ANC}.

These approaches have significant drawbacks.
First, the accelerometer signals do not always have the true motion artifacts (for example, finger tapping, wrist-twisting, and fist clenching/unfolding)~\cite{JBHI_FSM}, which limits the performance of the solutions heavily.
Second, the dominant frequencies from the accelerometer signals may overlap with the true frequencies of the HR, which makes signals impossible to differentiate by investigating the frequency domain.
Finally, since these solutions use multiple modalities~\cite{KID} to increase performance, they are more susceptible to sensor faults, such as communication or hardware failures.
Also, it is known that low-cost IMUs are vulnerable to various faults because of temperature variation, performance degradation, component damage~\cite{IMU_sensor_fault_1, IMU_sensor_fault_2}. 
The inaccurate information provided by the IMU will seriously affect the performance of these algorithms. 
Deep-learning solutions have been proposed to make the system more robust by only utilizing PPG signals~\cite{DeepPPG, PPGnet, Only_LSTM, Binary_CorNet}. 
However, as learning models have no information about the degree of motion artifact contamination in PPG signals, they can estimate HR with large errors. 
Or even though the IMU is fed together with PPG, the models cannot learn the relation between motion artifact and PPG signals and estimate the HR incorrectly~\cite{Benini_unprincipled},\textcolor{black}{\cite{Burrello_bs2}}. 
To prevent this, several works have proposed applying additional post-processing that considers the estimated values from preceding segments to detect and fix the abrupt changes in the model's output~\cite{Benini_unprincipled, DeepPPG}. 
Although current solutions have their own specific values, they remain straightforward and have not dealt with the rich dynamics of the cardiovascular system, which limits performance and overshadows the learning capability of algorithms. 
\\
In this work, we introduce a method that requires a single modality, PPG, to estimate the heart rate while concerning the temporal dynamics without requiring any post-processing. 
\section{Motivation}
\label{sec:motivation}
Several studies have attempted to model cardiovascular dynamics in heart rate estimation using linear methods such as mean filters~\cite{DeepPPG, Benini_unprincipled}, curve tracking~\cite{Curtoss} or linear regression models~\cite{temko}. 
Although the linear relation between temporal heartbeats can be useful to predict the current HR, the oscillations of a heart are complex and non-linear~\cite{complex_hr} while exhibiting a chaotic non-Markovian dynamical system~\cite{chaos_is_a_ladder}.

We first study the non-linear temporal relations of cardiovascular dynamics.
To investigate the relationship between heartbeat values over time, we calculated the Mutual Information (MI) between them.
Unlike linear methods such as correlation and mean filters, mutual information, which is a Shannon entropy-based measure of dependence\textcolor{black}{ ~\cite{shannon}} as defined in Equation~\ref{eq:Shannon} for variables $X$ and $Z$, captures non-linear statistical dependencies between variables and thus can act as a measure of true dependence~\cite{True_MI}.
\begin{equation}\label{eq:Shannon}
    I(X; Z) = \int_{\mathcal{X} \times \mathcal{Z}} \log\frac{d \mathbb{P}_{XZ}}{d\mathbb{P}_X \otimes \mathbb{P}_Z} \,d\mathbb{P}_{XZ} ,
\end{equation}
where $\mathbb{P}_{XZ}$ is the joint probability distribution, $\mathbb{P}_X = \int_Z \,d\mathbb{P}_{XZ}$ and $\mathbb{P}_Z = \int_{X} \,d\mathbb{P}_{XZ}$ are the marginals.
The mutual information between $X$ and $Z$ can be understood as the decrease of the uncertainty in $X$ given $Z$. 
We, therefore, calculated the mutual information between previous (past) heartbeats and the current one as in Equation~\ref{eq:MI}.
\begin{equation}\label{eq:MI}
    I(Y;\xi) := H(Y)-H(Y \hspace{0.5mm} | \hspace{1mm} \xi) ,
\end{equation}
where $\mathit{H}$, $Y$, and $\xi$ are the Shannon entropy, current heart rate, and past HR values, respectively.
$H(Y \hspace{0.5mm} | \hspace{0.5mm} \xi)$  is the conditional entropy of $Y$ given $\xi$. $I(Y; \xi)$ measures how much information on average previous heart rates convey about the current value without seeking any linear relations.

\begin{figure}[t]
    \centering
        \includegraphics[width=\columnwidth]{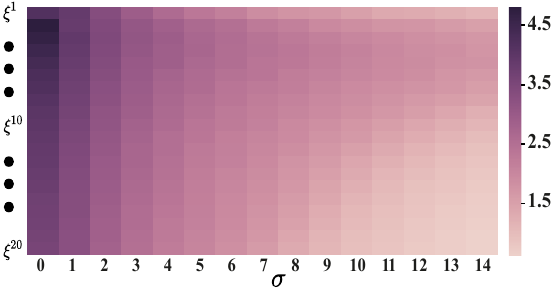}
    \caption{  \label{fig:MI} The heatmap of estimated mutual information between random variables $\boldsymbol{\xi}$ (previous heart rate values) and $Y$ (the current heart rate).
    We investigated the effect of time (y-axis) and noise (x-axis) for $\boldsymbol{\xi}$ with  $\boldsymbol{\xi} = \xi + \sigma \odot \epsilon$ where $\epsilon \sim \mathcal{N}(0, I)$. And, $\boldsymbol{\xi}^N = [y_{t-1},y_{t-2}, \dots, y_{t-N}]$.}
\end{figure}

Figure~\ref{fig:MI} shows that knowing previous HR values reduces the uncertainty of the current HR.
Moreover, it also indicates that even the past values are observed with an error, they carry information about the future.
Also, the noise affects the MI in different degrees depending on how many past values are included in $\boldsymbol{\xi}$.
The experiment for additive noise is especially important as it explains the effect from a statistical point of the possible realizations of past values on future estimations.
We note that small decreases in MI with longer windows come from finite-sample effects and estimator bias/variance, not a violation of the data processing inequality.

\subsection{The non-linearity}
Previous studies have often relied on leveraging prior HR values to address errors in predictions. 
However, many of these techniques were characterized by their straightforward and predominantly linear nature. 
Here, we seek to find any linear relations in HR using the Pearson correlation coefficient (PCC).
PCC calculates the linear correlation between two variables and ignores other types of relationships since it is defined as with covariance itself.

\begin{figure}[t]
    \centering
    \includegraphics[width=\columnwidth]{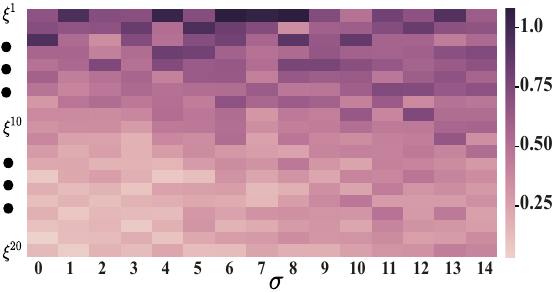}
    \caption{  \label{fig:PCC} The heatmap of calculated Pearson correlation between random variables $\boldsymbol{\xi}$ (previous heart rate values) and $Y$ (the current heart rate).
    We investigated the effect of time (y-axis) and noise (x-axis) for $\boldsymbol{\xi}$ with  $\boldsymbol{\xi} = \xi + \sigma \odot \epsilon$ where $\epsilon \sim \mathcal{N}(0, I)$. And, $\boldsymbol{\xi}^N = [y_{t}, y_{t-1},y_{t-2}, \dots, y_{t-N}]$.}
\end{figure}

When we calculate the PCC, we follow the same setup with MI estimation and take one variable as time and sum the absolute value of PCC instead of $\pm$\,1 since the overall trend is important rather than increase and decrease.
Figure~\ref{fig:PCC} shows the PCC values where no clear and distinct pattern is observed at all.
Conversely, there are some cases in which the addition of noise increases the correlation between past and future HRs.
These experiments show that HR changes are non-linear.
Thus, we propose a method to capture non-linear variations for improved HR estimation, unlike the linear methods used in previous works~\cite{DeepPPG, Benini_unprincipled, temko}.
\section {Objective}
\label{sec:Prop_method}
Considering our findings, we aim to train a learner $f_{\theta}$ which seeks to learn the behavior of heart rate variability with volumetric changes in blood together for estimating the current heart rate value. 
Therefore, we only require volumetric blood volume change signals to estimate the heart rate, unlike the related works which require additional modalities~\cite{Curtoss, Sarkar_Etemad_2021}.

\subsection{Framework}

The overall proposed framework can be explained as follows: 
Given an observation of the blood volume signal $\boldsymbol{\mathrm{x_t}}$ that is a single dimensional vector drawn from the distribution $\mathcal{X}$ at time $t$ and composed of a sequence of voltage values with length T, $(x_1, x_2, \dots, x_T)$, we aim to map this sequence to the heart rate value of this segment $y_t$.

Previous deep learning approaches attempted to estimate the conditional probability $p(y_t|\boldsymbol{\mathrm{x_t}})$ by training different mapping functions, using Equation~\ref{eq:loss_previous}, that perform $f_{\theta}: \mathcal{X} \rightarrow \mathcal{Y}$ where the $\mathcal{Y}$ is a one-dimensional space consisting of real-valued heart rate $y \in \mathcal{Y}$, and $f_{\theta}$ is a parametric family of mappings (e.g., CNNs), under a loss function $L$. 
\begin{equation}\label{eq:loss_previous}
    \min_{\theta} \sum_t L (f_{\theta}(\boldsymbol{\mathrm{x_t}}), y_t)
\end{equation}
As the previous methods optimized the models to estimate $y_t$ using only the current signal $\boldsymbol{\mathrm{x_t}}$, they lack the ability to capture heart rate variability. 
As a result, when inputs deviate from distributions seen during training due to noise, the models produce large errors~\cite{Benini_unprincipled, DeepPPG}, decreasing the performance.

To solve this problem and make the trained model suitable for real-life conditions, we train the model with the current signal and HR variations together.
We optimize the model to estimate the heart rate distribution given the volumetric changes in blood and HR variations, i.e.,  $p(y_t|\boldsymbol{\mathrm{x_t}},\boldsymbol{\xi_t})$ where $y_t$ is the heart rate of the current segment and $\boldsymbol{\xi_t}$ is the vector composed of previously estimated HR values.

We optimized the training task for a set of input-target pairs using the loss function shown in Equation~\ref{eq:optimization} where the network function $f_{\theta}$, is parameterized by $\theta$:
\begin{equation}\label{eq:optimization}
    \mathcal{L}_{\text{MAE}}(\theta) = \mathbb{E} [ ||y_t - f_{\theta}(\boldsymbol{\mathrm{x_t}},y_{t-1},\dots, y_{t-K}) ||_1]
\end{equation}
The loss function employed to train the learner is closely similar to transfer entropy, with the goal of maximizing it.
Transfer entropy, $T_{Y \rightarrow X}$, is a non-parametric statistic measuring the rate of information flow between two random variables~\cite{causal}.
\begin{equation}
\sum p(y_t, y_{t-L}, x_{t-L}) \log\left(\frac{p(y_t|y_{t-L}, x_{t-L})}{p(y_t|y_{t-L})}\right)
\end{equation}
In our case, we train the model to learn the information flow from the past to the future while incorporating the present.
Hence, we characterize this as the conditional mutual information between variables, as defined in Equation~\ref{eq:cMI}.
\begin{equation}\label{eq:cMI}
    T_{Y \rightarrow X} = I(X_t;Y_{t} | X_{t-1:t-L})
\end{equation}
Learning the function $f_{\theta}$ by minimizing Equation~\ref{eq:ERM}---also known as the empirical risk minimization---maximizes the conditional mutual information between the past, present, and future states of the dynamical complex system.
\begin{equation}\label{eq:ERM}
    \min_{\theta} \sum_t L (f_{\theta}(\boldsymbol{\mathrm{x_t}},\boldsymbol{\xi_t}), y_t)
\end{equation}
In contrast to common dynamical systems~\cite{markovian_ICML,NIPS_markovian}, cardiovascular dynamics are not Markovian due to their chaotic nature, such as we have no information when someone starts exercising and how it will affect the heart. 
We, therefore, incorporate our findings from MI analysis into a common network and show how a small modification can improve performance significantly.
This non-linear relationship, as revealed through MI analysis (shown in Figure~\ref{fig:MI},~\textcolor{black}{derived from Equations~1 and 2}), is instrumental in enhancing the model's performance.

\subsection{Architecture}
\begin{figure}[t]
    \centering
    \includegraphics[width=1\columnwidth]{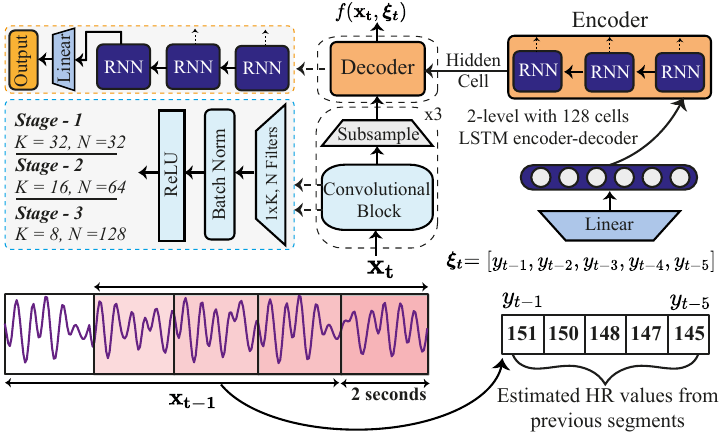}
    \caption{Modified encoder-decoder architecture. 
    The encoder part incorporates the HR variations into the model. 
    The decoder outputs the HR by using both the current signal and encoded HR. 
    Black arrows represent the input-output flow while double arrows show the content of a block.}
    \label{fig:model}
\end{figure}

We modify the encoder-decoder architecture to learn heart rate variations together with blood volume changes.
Specifically, we used linear layers with Long-Short Term Memory (LSTM) to encode the previous HR estimations and CNNs with LSTMs to extract features from the current segment, which is common in literature for sequential data~\cite{Only_LSTM}, as shown in Figure~\ref{fig:model}. 
We first expand the HR values using a linear transformation with fifty output features. 
Then, extracted features are fed to the encoder.
Our feature extractor includes convolutional blocks that are composed of 1D convolutions followed by batch normalization~\cite{batch} and ReLU. 
We added a max-pooling layer with a size of two after each convolutional block.
The decoder takes the extracted features from the output of the last pooling layer and conditions the encoder output. 
To estimate the HR, we used the last hidden state of the decoder with a linear layer.
We also used a data augmentation to increase the diversity of heart rate variations. 
We, therefore, added a random noise sampled from the normal distribution to the HR, which is given in Equation~\ref{eq:gaussian}. 
At the end of augmentation, some segments that are chosen randomly map to the different heart rate vectors $\boldsymbol{\xi}$, with small differences. 
\begin{equation}\label{eq:gaussian}
    \xi_{aug} = \boldsymbol{\xi}  + 3 \cdot \boldsymbol{z} \hspace{2mm} \text{where $\boldsymbol{z} \sim \mathcal{N}(0, 1)$ } 
\end{equation}
Although this procedure adds noise during training, we observed significant performance gains, especially on smaller datasets, as it exposes the model to HR variations not present in the training data.  
\textcolor{black}{Importantly, the added perturbations are small relative to overall HR dynamics.  
For example, in a commonly used large PPG dataset~\cite{DeepPPG}, HR values have a standard deviation of $\approx 17\,$\,bpm.  
With added noise of $\sigma$=3\,bpm, the resulting Signal-to-Noise ratio (SNR) is $(17/3)^2 \approx 32$, meaning most of the original HR information is preserved.}  
We randomly selected 10\% of the HR vectors in the training data (excluding validation and test) for this augmentation.

\section{Experimental Setup}
\label{sec:Exp_setup}	

\begin{table*}[h]
\centering
\caption{Performance comparison of methods.  \textcolor{black}{Best results are shown in \textbf{bold}, and second-best are \underline{underlined}}. }
\label{tab:performance}
\renewcommand{\arraystretch}{1}
\setlength{\tabcolsep}{8pt}
\begin{adjustbox}{width=2\columnwidth,center}
\begin{threeparttable}
\begin{tabular}{@{}lllllll@{}}
\toprule
\multicolumn{1}{l}{\begin{tabular}[l]{@{}l@{}}\textbf{Work (Year)} \end{tabular}}& \multicolumn{1}{l}{\begin{tabular}[l]{@{}l@{}} \textbf{Input} \end{tabular}}& \multicolumn{1}{c}{\begin{tabular}[l]{@{}l@{}} \textbf{Pre-Processing}\end{tabular}} & \multicolumn{1}{l}{\begin{tabular}[l]{@{}l@{}}\textbf{Method} (\# of params)\end{tabular}} & \multicolumn{1}{l}{\begin{tabular}[l]{@{}l@{}} \textbf{Post-Processing} \end{tabular}} & \multicolumn{1}{l}{\begin{tabular}[l]{@{}l@{}} \textbf{Dataset} \end{tabular}} & \multicolumn{1}{l}{\begin{tabular}[l]{@{}l@{}} \textbf{Results (bpm)} \end{tabular}}\\ \midrule
\textit{Heuristic Solutions} \\ \midrule

\multirow{1}{*}{Troika${\tnote{*}}$~\cite{TROIKA}} &  \multirow{1}{*}{PPG + ACC} & \multirow{1}{*}{\begin{tabular}[l]{@{}l@{}}Signal decomposition \end{tabular}} & \multirow{1}{*}{\begin{tabular}[l]{@{}l@{}}Signal reconstruction\end{tabular}} & \multirow{1}{*}{N/A} & SPC15-12 & 2.34 \\ \midrule

 \multirow{4}{*}{SpaMa${\tnote{**}}$ ~\cite{DeepPPG}}&  \multirow{4}{*}{PPG + ACC} & \multirow{4}{*}{\begin{tabular}[l]{@{}l@{}}0.5- to 3-Hz filtering\\ Downsampling\\ Freq. Transformation\end{tabular}} & \multirow{4}{*}{\begin{tabular}[l]{@{}l@{}}Spectral filtering using\\ power spectrum\end{tabular}} & \multirow{4}{*}{Interpolation} & SPC15-12 & 13.1$\pm$20.7 \\
 &   &  &  & &SPC15-22 &9.20$\pm$11.4 \\
 &  &  &  &  &Dalia & 15.56$\pm$7.5 \\
 &  &  &  &  &WESAD & 11.51$\pm$3.7 \\ \midrule

\multirow{4}{*}{SpaMaPlus~\cite{DeepPPG}}& \multirow{4}{*}{PPG + ACC} & \multirow{4}{*}{\begin{tabular}[l]{@{}l@{}}0.5- to 3-Hz filtering\\ Downsampling\\ Freq. Transformation\end{tabular}} & \multirow{4}{*}{\begin{tabular}[l]{@{}l@{}}Picking the maximum \\ in a weighted spectrum\\  using linear prediction\end{tabular}} & \multirow{4}{*}{HR tracking} & SPC15-12 & 4.25$\pm$5.9 \\
 &   &  &  &  & SPC15-22 & 12.31$\pm$15.5 \\
 &   &  &  &  &Dalia & 11.06$\pm$4.8 \\ 
 &   &  &  &  &WESAD & 9.45$\pm$2.9 \\ \midrule
 
\multirow{4}{*}{\begin{tabular}[l]{@{}l@{}}Schaeck~\cite{DeepPPG}\\\cite{sch_original}\end{tabular}} &  \multirow{4}{*}{PPG + ACC} & \multirow{4}{*}{\begin{tabular}[l]{@{}l@{}}0.5- to 6-Hz filtering\\ Downsampling\\ Freq. Transformation\end{tabular}} & \multirow{4}{*}{\begin{tabular}[l]{@{}l@{}}Comparing spectra of \\ PPG and accelerometer\end{tabular}} & \multirow{4}{*}{HR tracking} &SPC15-12 & 2.91$\pm$4.6 \\
 &   &  &  & &SPC15-22&24.65$\pm$24 \\
 &  &  &  &  &Dalia & 20.45$\pm$7.1 \\
 &  &  &  &  &WESAD & 19.97$\pm$8.1 \\\midrule
 

\multirow{4}{*}{CurToSS${\tnote{*}}$~\cite{Curtoss}} & \multirow{4}{*}{PPG + ACC} & \multirow{4}{*}{\begin{tabular}[l]{@{}l@{}} Sparse spectrum\\ reconstruction \end{tabular}} & \multirow{4}{*}{Curve tracing} & \multirow{4}{*}{N/A} & SPC15-12 & 2.2$\pm$-- \\
&  &  &  & &SPC15-22&4.5$\pm$-- \\
& &  &  &  & Dalia & 5.0$\pm$2.8 \\
& &  &  &  & WESAD & 6.4$\pm$1.8 \\ \midrule
\textit{Deep Learning Solutions} \\ \midrule
\multirow{2}{*}{CNN~\cite{DeepPPG}} & \multirow{2}{*}{PPG + ACC} & \multirow{2}{*}{\begin{tabular}[l]{@{}l@{}}0- to 4-Hz filtering\\ Freq. Transformation\end{tabular}} & \multirow{2}{*}{CNN ensemble (\textit{8.5M})} & \multirow{2}{*}{N/A} & SPC15-12 & 4.0$\pm$5.4 \\
 &   &  &  &  & SPC15-22 &16.51$\pm$16.1 \\ \midrule
 \multirow{2}{*}{PPGNet~\cite{PPGnet}} &  \multirow{2}{*}{PPG} & \multirow{2}{*}{\begin{tabular}[l]{@{}l@{}}0.4- to 18-Hz filtering\\ z-score normalization\end{tabular}} & \multirow{2}{*}{CNN + LSTM (\textit{765k})} & \multirow{2}{*}{N/A} & SPC15-12 & \underline{3.36} \\
 &   &  &  &  & SPC15-22 & 12.48 \\ \midrule
 \multirow{2}{*}{\begin{tabular}[l]{@{}l@{}}CorNet~\cite{CorNET}\\\cite{Binary_CorNet}\end{tabular}} &  \multirow{2}{*}{PPG} & \multirow{2}{*}{\begin{tabular}[c]{@{}c@{}}0.4- to 18-Hz filtering\\ z-score normalization\end{tabular}} & \multirow{2}{*}{CNN + LSTM (\textit{257k})} & \multirow{2}{*}{N/A} & SPC15-12 & 4.67 \\
 &  &  &  &  & SPC15-22 &  \underline{5.55} \\ \midrule
\multirow{2}{*}{BinCorNet~\cite{Binary_CorNet}} & \multirow{2}{*}{PPG} & \multirow{2}{*}{\begin{tabular}[l]{@{}l@{}}0.4- to 18-Hz filtering\\ z-score normalization\end{tabular}} & \multirow{2}{*}{CNN + LSTM} & \multirow{2}{*}{N/A} & SPC15-12 & 6.78 \\
 &  &  &  &  &  SPC15-22 & 7.32 \\ \midrule
 \multirow{2}{*}{CardioGAN~\cite{Sarkar_Etemad_2021} } & \multirow{2}{*}{PPG} & \multirow{2}{*}{\begin{tabular}[l]{@{}l@{}}1- to 8-Hz filtering\\ z-score normalization \end{tabular}} & \multirow{2}{*}{GANs  (\textit{5.1M})} & \multirow{2}{*}{N/A} & Dalia & 8.4 \\
 &  &  &  &  & WESAD & 8.6 \\ \midrule
 
\multirow{1}{*}{Only-LSTM~\cite{Only_LSTM}} & \multirow{1}{*}{PPG} & \multirow{1}{*}{\begin{tabular}[l]{@{}l@{}} N/A \end{tabular}} & \multirow{1}{*}{LSTM (\textit{680k})} & \multirow{1}{*}{N/A} & SPC15-12 & 4.47$\pm$3.68\\ \midrule

\multirow{1}{*}{TimePPG~\cite{Benini_unprincipled}} & \multirow{1}{*}{PPG + ACC} & \multirow{1}{*}{\begin{tabular}[l]{@{}l@{}} 0.5- to 4-Hz filtering \end{tabular}} & \multirow{1}{*}{TCN (\textit{900k})} & \multirow{1}{*}{HR clipping} & Dalia & 4.88$\pm$3.23 \\ \midrule

\multirow{1}{*}{LSTM~\cite{Newest}} & \multirow{1}{*}{PPG + ACC} & \multirow{1}{*}{\begin{tabular}[l]{@{}l@{}} N/A \end{tabular}} & \multirow{1}{*}{LSTM} & \multirow{1}{*}{Filtering} & Dalia & 7.44$\pm$3.26 \\ \midrule

\textcolor{black}{\multirow{1}{*}{KID${\tnote{**}}$\hspace{2mm}~\cite{KID} } } & \multirow{1}{*}{PPG + ACC} & \multirow{1}{*}{\begin{tabular}[l]{@{}l@{}} Adaptive linear filtering \end{tabular}} & \multirow{1}{*}{CNN + Attention} & \multirow{1}{*}{N/A} & Dalia & \textbf{3.79}\\ \midrule

\multirow{1}{*}{WildPPG} & \multirow{1}{*}{PPG} & \multirow{1}{*}{\begin{tabular}[l]{@{}l@{}} 0.5- to 4-Hz filtering \end{tabular}} & 1D ResNet & N/A & WildPPG & \underline{8.62}$\pm$0.06 \\ \midrule
\multirow{4}{*}{\textbf{Ours (2025)}} &  \multirow{4}{*}{PPG} & \multirow{4}{*}{\begin{tabular}[l]{@{}l@{}}0.5- to 4-Hz filtering\\z-score normalization \end{tabular}} & \multirow{4}{*}{\begin{tabular}{@{}l@{}}Encoder-Decoder\\ \makebox[1cm][r]{+} \hspace{3mm}(\textit{430k}) \\ \hspace{1mm} HR variations \end{tabular}} & \multirow{4}{*}{N/A} & SPC15-12 & \hspace{-4mm} \textbf{2.24}\scriptsize{$\pm$1.03 \color{Green}{(+33\%)}} \\
 &  &  &  & & SPC15-22& \hspace{-4mm} \textbf{4.30}\scriptsize{$\pm$2.97 \color{Green}{(+22\%)} }  \\
 & &  &  &  & Dalia & \hspace{-4mm} \underline{4.16}\scriptsize{$\pm$1.74 \color{purple}{(-10\%)}} \\
 & &  &  &  & WildPPG & \hspace{-4mm} \textbf{7.73}\scriptsize{$\pm$0.05 \color{Green}{(+12\%)}} \\
 & &  &  &  & WESAD & \hspace{-4mm} \textbf{4.75}\scriptsize{$\pm$2.25 \color{Green}{(+45\%)}} \\ \bottomrule
\end{tabular}
\begin{tablenotes}
  \item [*]  \normalsize Dataset-specific threshold optimization is performed, requiring prior knowledge of HR distribution and activity intensity.
  \item [**] \normalsize \textcolor{black}{Increasing dataset size with generating synthetic PPG waveforms to form the high HR signals using speed up~\cite{speed_up}.}
  \end{tablenotes}
  \end{threeparttable}
  \end{adjustbox}
\end{table*}

\subsection{Datasets}
\label{sec:Datasets}
We used four popular PPG datasets: IEEE Signal Processing Cup 2015 (SPC15)~\cite{IEEE_SPC2, IEEE_SPC_2015}, Dalia~\cite{DeepPPG}, WildPPG~\cite{wildppg}, and WESAD~\cite{WESAD}, which are selected to assess performance across diverse activities and age groups.

\textit{IEEE SPC:} Includes 22 recordings from participants aged 18–58~\cite{Binary_CorNet}.
Data includes accelerometer signals, two PPG signals, and ECG sampled at 125 Hz, with sensors worn on the wrist. 
Ground truth ECG was recorded via a chest sensor.

\textit{Dalia:} Recorded from 15 participants, with each session lasting 2 hours~\cite{DeepPPG}.
ECG and PPG signals were collected during daily activities such as sitting, walking, and cycling.

\textit{WildPPG:} Contains 13.5 hours of recordings from 16 participants during outdoor activities~\cite{wildppg}.
It captures heart rate data under challenging conditions with extreme noisy environments, including temperature variations.

\textit{WESAD:} Includes 15 participants (12 males, 3 females, mean age 27.5)~\cite{WESAD}, with over one hour of recordings during tasks like solving arithmetic problems and watching videos.

\subsection{Evaluation}
Our evaluation uses leave-one-session-out (LOSO) cross-validation (CV).
In LOSO, each test session is excluded from training, allowing assessment under realistic conditions with subject variations.
Then, a 3-fold CV is used on the training set to select the model with the lowest validation error. 
During testing, we generated the $\boldsymbol{\xi}$ using model predictions without any information about the test subject.
For the first four segments, HR was estimated using Fourier transformation by selecting the harmonic with the highest spectral energy.
\\
We evaluated performance using the most common metric in prior work, the mean absolute error (MAE)~\cite{DeepPPG}.

\subsection{Training Architecture}
\label{train-CNN}
The designed architecture takes as input only the pre-processed PPG and no other patient- or PPG-related features. 
We used the Adam optimizer~\cite{Adam} with $\beta_1=0.9$, $\beta_2=0.999$, and a mini-batch size of 32. 
The learning rate was initialized to $5e-4$ with a weight decay of $1e-6$ and reduced by $0.1$ when the validation loss stopped improving for 10 consecutive epochs. 
The training continues until 30 successive epochs without validation performance improvements. 
The best model is chosen as the lowest mean absolute error on the validation. 


\section{Experimental Results and Analysis}
\label{sec:Results}	
We evaluated our method with fixed architecture and hyperparameters on four datasets.
Table~\ref{tab:performance} presents the results, along with the percentage performance gain of our method compared to the best learning-based solution for each dataset.
We compared with the learning-based algorithms because traditional rule-based methods have optimized their parameters and thresholds according to the dataset statistics, which require prior knowledge of HR behavior with activities.

Our method outperforms all related works while using fewer sensor modalities.
Many published approaches rely on additional inputs, such as power spectra, raw signals, or their combination with multiple sensor modalities, which makes their models more complex.
Yet, their performance remains lower than ours, showing that our method is more effective for real-world settings where data is heavily contaminated with noise.
\textcolor{black}{We also examined model performance across demographic and activity-related factors.  
While no clear differences were observed across age or sex groups, our HR-integrated architecture showed greater improvements during high-motion activities. 
In the Dalia, which includes activity labels, our method reduced error by $\approx$ 10--15\% in \textit{stairs} and \textit{table soccer} segments, where error is typically high~\cite{DeepPPG}.
}

Table~\ref{tab:performance} shows that the performance of related works, which evaluated their method on multiple datasets, vary significantly between datasets since the Dalia and SPC15-22 datasets are seen as the most challenging to estimate HR due to including a wide range of activities performed under close to real-life conditions~\cite{DeepPPG}. 
While these datasets are challenging due to low signal quality, our method reduces overall error and either outperforms or \textcolor{black}{closely matches prior methods (e.g., second-best on Dalia)}.
The results indicate that the performance of models should be evaluated across datasets with the same hyperparameters to investigate if the models generalize and perform well enough under numerous types of activities with different noise levels.
Otherwise, the architectures and hyperparameters (post-processing with threshold values) can be optimized for a dataset while performing poorly for the rest.

Overall, our method improves HR estimation by 30--45\% compared to prior solutions.
This shows that learning HR variations together with volumetric flow changes help the model to accurately estimate heart rate under real-life settings.
\begin{figure*}[t]
    \centering
    \includegraphics[width=\linewidth]{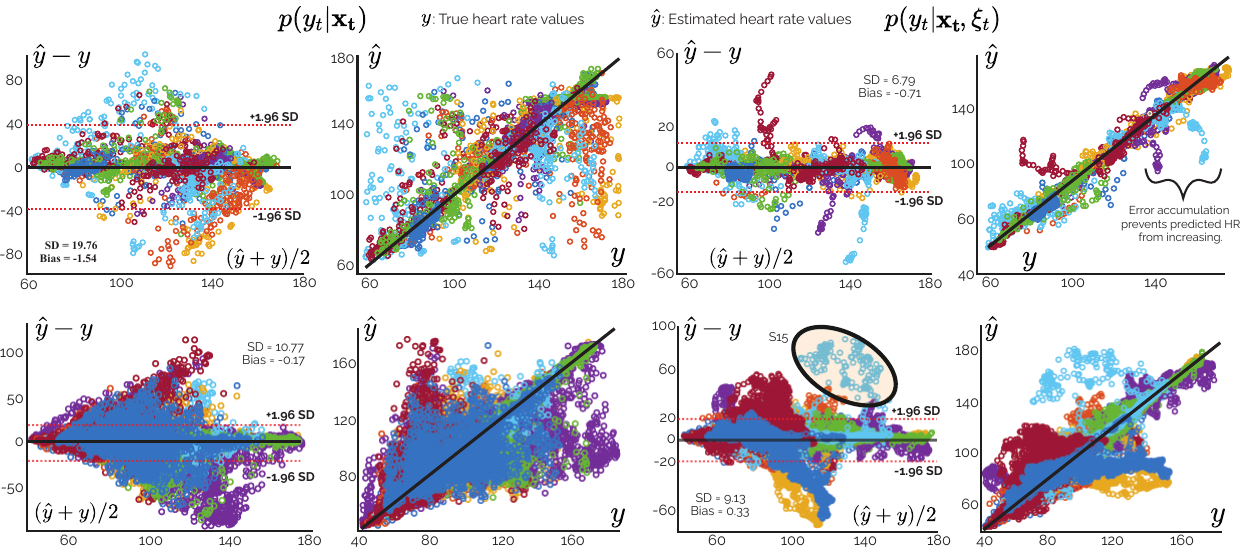}
    \caption{Bland-Altman and correlation plots.
    Top row: SPC15-22, bottom row: Dalia.
    Left column: without learning the heart rate variations, right column: learning variations with $\boldsymbol{\xi}$ vector for the HR estimation.
    Colors represent subjects.}
    \label{fig:bland-altman}
\end{figure*}
\vspace{-5mm}
\subsection{Ablation Study}
\label{sec:ablation}
\begin{table}[t]
\centering
\caption{Ablation studies across datasets. Top: model improvements with HR dynamics and augmentation. Bottom: comparison of different $\xi$ lengths for past HR inputs.}
\begin{adjustbox}{width=1\columnwidth,center}
\label{tab:combined_ablation}
\renewcommand{\arraystretch}{1.1}
\begin{tabular}{@{}lccccc@{}} 
\textbf{Method / Datasets} & SPC15-12 & SPC15-22 & Dalia & WESAD & WildPPG \\ \midrule
\multicolumn{6}{l}{\textit{Model Variants}} \\
Traditional, $p(y_t|\boldsymbol{x_t})$ & 7.80$\pm$4.71 & 11.11$\pm$6.21 & 8.10$\pm$5.45 & 6.13$\pm$3.52 & 8.62$\pm$0.06 \\
+ HR variations & 5.85$\pm$3.31 & 6.50$\pm$5.15 & 6.12$\pm$3.20 & 4.88$\pm$2.45 & 7.92$\pm$0.02 \\
+ Data Aug. \textbf{(Proposed)} & \textbf{2.24}$\pm$1.03 & \textbf{4.30}$\pm$2.97 & \textbf{4.16}$\pm$1.74 & \textbf{4.75}$\pm$2.25 & \textbf{7.73}$\pm$0.05 \\
\midrule
\multicolumn{6}{l}{\textit{Effect of $\xi$ Length (Past HR Window)}} \\
$\xi_3 = y_{t-1}, ..., y_{t-3}$ & 3.37$\pm$1.69 & 5.23$\pm$3.14 & 5.68$\pm$2.65 & 5.34$\pm$2.57 & 8.28$\pm$0.10 \\
$\xi_5 = y_{t-1}, ..., y_{t-5}$ & \textbf{2.24}$\pm$1.03 & \textbf{4.30}$\pm$2.97 & \textbf{4.16}$\pm$1.74 & \textbf{4.75}$\pm$2.25 & \textbf{7.73}$\pm$0.05 \\
$\xi_7 = y_{t-1}, ..., y_{t-7}$ & 3.35$\pm$1.13 & 5.19$\pm$3.11 & 5.62$\pm$2.28 & 5.14$\pm$2.35 & 7.79$\pm$0.17 \\
\bottomrule
\end{tabular}
\end{adjustbox}
\end{table}
To investigate the effectiveness of our method, we perform ablations by removing each of the components individually. 
For the \textit{Traditional} case, we only feed the signal similar to previous solutions.
For the second part, we used the complete model, which learns HR variations, without augmentations.

Table~\ref{tab:combined_ablation} shows that incorporating our method improves estimation.
Notably, augmentation reduces error for the smaller datasets, SPC15 pairs, more than in the bigger dataset, Dalia.
This evidence supports our motivation for introducing HR variations to the model.
Interestingly, increasing temporal dependencies with longer \( \xi \) reduces performance, despite providing more information for learning non-linear HR variations.
\begin{table}[b]
\centering
\caption{\textcolor{black}{Architecture ablations on WildPPG. Values in parentheses indicate baselines without learning HR dynamics.}}
\begin{adjustbox}{width=1\columnwidth,center}
\label{tab:architecture}
\renewcommand{\arraystretch}{1}
\begin{tabular}{@{}lccccc@{}} 
\textcolor{black}{\textbf{Dataset / Models}} & \textcolor{black}{CNN~\cite{DeepPPG}} & \textcolor{black}{LSTM~\cite{Only_LSTM}} & \textcolor{black}{ CNN+Attention~\cite{KID} } &  \textcolor{black}{Ours} \\ \midrule
\textcolor{black}{WildPPG} & \textcolor{black}{7.80 (8.87)} & \textcolor{black}{7.98 (9.03)} & \textcolor{black}{ 7.59 (8.53)} &  \textcolor{black}{7.73 (8.62) } \\
\bottomrule
\end{tabular}
\end{adjustbox}
\end{table}
In other words, the model performs better when past HRs are included to a certain extent. 
This is an interesting finding, revealing that incorporating past information into the model is effective up to a limit. 
And, inputting all data into the model in hopes of uncovering relationships decreases the performance.
\textcolor{black}{
To further understand the model's temporal behavior, we also analyzed its performance during sudden HR changes.
In the Dalia dataset, the largest HR jump between consecutive segments was 16.46 bpm.
For such high-change intervals, the model’s average error is 4.52 bpm—around 10\% higher than the dataset average.
This suggests that while past information is helpful, handling abrupt HR transitions may require modifications, such as dynamic modeling or specialized loss terms.
}
\textcolor{black}{
Additionally, we performed ablations whether learning HR dynamics is architecture-agnostic and improves performance across models with minimal changes.  
Specifically, we used a two-layer MLP to extract features from past HR values and concatenated them with PPG features before the final prediction layer.  
Results in Table~\ref{tab:architecture} show that our method improves performance across all models.
}
We, also, show the Bland-Altman plot with correlation for SPC and Dalia in Figure~\ref{fig:bland-altman} to investigate how learning the HR variability affects estimations.
When the model takes only the current segment as input, it produces large errors.
However, if the model learns the HR variations besides the current segment, the error decreases.


\section{Discussion and Limitations}
We propose a HR estimation algorithm that leverages temporal dynamics to enhance cardiac monitoring with wearable devices, particularly in real-life conditions where noise is prevalent.
Our method surpasses current state-of-the-art approaches by reducing estimation errors by up to 40\%, demonstrating improvements in robustness and accuracy.

\paragraph{Comparison with previous methods}
Unlike traditional methods that rely heavily on heuristic-based signal processing techniques, our approach integrates non-linear temporal dynamics.
While previous works~\cite{DeepPPG} focused on hand-crafted features or trial-and-error optimization in deep learning models, our method models the chaotic behavior of cardiovascular dynamics, uncovering temporal patterns.
This leads to more stable HR estimations, even under motion-corrupted conditions where conventional algorithms fail.
Moreover, our method reduces dependency on multiple sensing modalities, outperforming multi-sensor approaches while using fewer modalities.
\textcolor{black}{
While our method ranks second on the Dalia, trailing KID---uses adaptive filtering with inertial data and attention-based models---by 10\%~\cite{KID}), our approach is complementary.  
In addition, our model is relatively lightweight and avoids reliance on inertial signals or attention mechanisms, making it more suitable for deployment on memory- and energy-constrained wearable devices, which is an important design factor for HR estimation~\cite{Burrello_bs1}.  
In future work, combining our dynamics module with other methods, including data augmentation~\cite{Burrello_bs3}, could further improve the performance.
}

\paragraph{Clinical Relevance}
Accurate HR estimation is critical in ambulatory settings, where continuous monitoring is essential for managing cardiovascular diseases~\cite{sandvik_heart_1995}, stress detection, and fitness tracking~\cite{HRV_UCI}.
By minimizing errors and reducing reliance on sensors, our method offers a solution for wearable devices.
The ability to maintain high accuracy in dynamic environments enhances the reliability of HR monitoring for early detection of abnormal heart rate patterns, contributing to preventive healthcare.

\paragraph{Limitations and Future Work}
Despite its strengths, our approach has limitations.
First, it relies on accurate initial HR values to predict subsequent readings, making it sensitive to noise during the initial segments.
In noisy scenarios, heuristic methods like Fourier transforms may struggle to extract reliable initial HR, potentially affecting the performance.

Second, our model requires tuning of the temporal window size to effectively utilize historical data. 
While we identified an optimal range, incorporating excessive historical information can degrade performance and increase computational complexity.
Future research can focus on adaptive mechanisms that adjust window sizes based on real-time signal quality \textcolor{black}{or detecting and handling the motion artifacts~\cite{handling_artifact}}.

Lastly, while our method shows strong performance on datasets, further validation in clinical settings with patients is necessary to confirm its generalizability.
\textcolor{black}{Testing on datasets that include abnormal heart rate would help evaluate the model's sensitivity to irregular and sudden changes in HR.}
\section {Conclusion}
\label{sec:Conclusion}
Our work is the first to study the behavior of HR variations through mutual information and have introduced an approach for improving HR estimations. 
Our method overcomes previous limitations by learning the HR dynamics together with blood volume changes to prevent large errors made by previously proposed learning methods.
As a result, our method works without requiring post-processing or optimizing hyperparameters, e.g., threshold values.
These findings underscore the efficacy of incorporating past information into the model, up to a certain threshold, while random inclusion of historical data decreases the performance while increasing computations.
To the best of our knowledge, our method achieves the best performance, decreasing the error rates up to 40\%, on HR estimation compared to the existing solutions.

\section*{References}
\vspace{-5mm}
\bibliographystyle{IEEEtran}
\bibliography{sample}

\end{document}